\pdfoutput=1

\documentclass[11pt]{article}

\usepackage[preprint]{acl}

\usepackage{times}
\usepackage{latexsym}

\usepackage[T1]{fontenc}

\usepackage[utf8]{inputenc}

\usepackage{microtype}
\microtypesetup{protrusion=true, expansion=true}

\usepackage{inconsolata}

\usepackage{graphicx}

\title{AutoIntent: AutoML for Text Classification}

\author{
 \textbf{Alekseev Ilya\textsuperscript{1,2,4}},
 \textbf{Solomatin Roman\textsuperscript{1,3}},
 \textbf{Rustamova Darina\textsuperscript{3}},
\\
 \textbf{Kuznetsov Denis\textsuperscript{1}}
\\
\\
 \textsuperscript{1}Moscow Center for Advanced Studies,
 \textsuperscript{2}Moscow State University,\\
 \textsuperscript{3}ITMO University,
 \textsuperscript{4}dresscode.ai.
\\
 \small{
   \textbf{Correspondence:} \href{mailto:ilya\_alekseev\_2016@list.ru}{ilya\_alekseev\_2016@list.ru}
 }
}

\usepackage{tikz}
\def\checkmark{\tikz\fill[scale=0.4](0,.35) -- (.25,0) -- (1,.7) -- (.25,.15) -- cycle;} 

\begin{document}
\maketitle
\begin{abstract}
AutoIntent is an automated machine learning tool for text classification tasks. Unlike existing solutions, AutoIntent offers end-to-end automation with embedding model selection, classifier optimization, and decision threshold tuning, all within a modular, sklearn-like interface. The framework is designed to support multi-label classification and out-of-scope detection. AutoIntent demonstrates superior performance compared to existing AutoML tools on standard intent classification datasets and enables users to balance effectiveness and resource consumption.
\end{abstract}

\renewcommand{\arraystretch}{1.2}  

\section{Introduction}
Text classification remains a fundamental task in natural language processing, with applications ranging from intent detection in conversational systems \cite{weld2021surveyjointintentdetection} to content categorization and sentiment analysis \cite{TAHA2024100664}. Modern NLP has been revolutionized by transformer-based embedding models \cite{vaswani2023attentionneed, devlin2019bertpretrainingdeepbidirectional, reimers-2019-sentence-bert}, which provide rich contextual representations of text. However, effectively utilizing these models for classification tasks requires careful consideration of multiple components: the choice of pre-trained embedding model, the selection of appropriate classification algorithms, and the optimization of their hyperparameters.

\begin{figure}[!htb]
  \centering
  \includegraphics[width=1\linewidth]{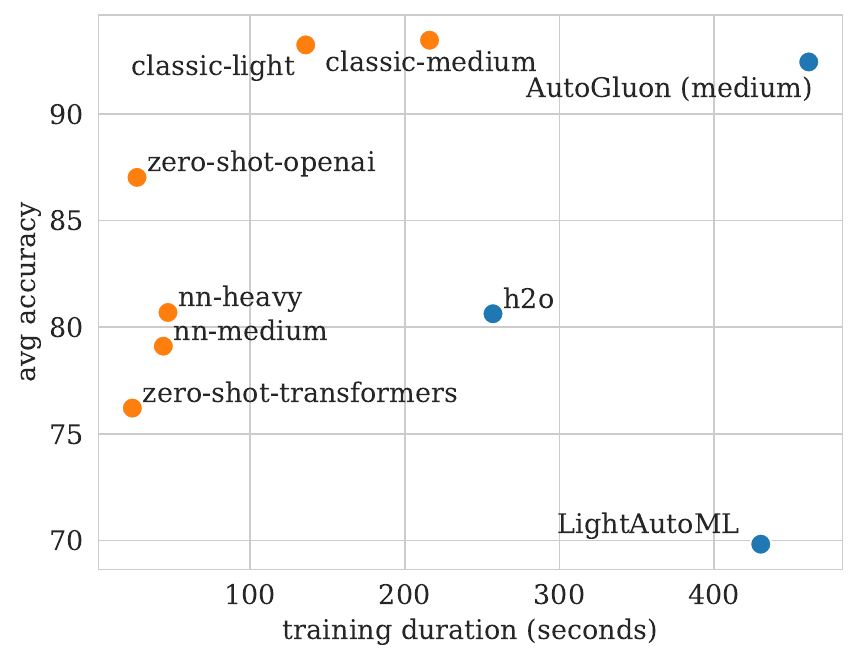}
  \caption{Average accuracy (banking77 \cite{banking77}, massive \cite{massive}, minds14 \cite{minds14}, hwu64 \cite{hwu64}) and training duration (on minds14) of AutoIntent presets (\textbf{orange}) and baseline AutoML tools (\textbf{blue}).}
  \label{fig:main_res}
\end{figure}

Traditional machine learning approaches often require manual tuning of these components, which can be time-consuming and requires significant expertise. While automated machine learning (AutoML) frameworks \cite{baratchi2024automated} have emerged to automate this process, existing solutions for NLP tasks often lack comprehensive support for the full spectrum of hyperparameter optimization (for instance, choosing best embedding model) and tuning confidence thresholds for handling multi-label classification and out-of-scope (OOS) detection \cite{clinc150}.

This paper introduces AutoIntent\footnote{\url{https://github.com/DeepPavlov/AutoIntent}}, a novel AutoML framework specifically designed for intent classification tasks. The framework offers a sklearn-like interface \cite{pedregosa2018scikitlearnmachinelearningpython} for ease of use and supports even multi-label classification and OOS detection, bridging the gap between AutoML and conversation systems.

\begin{table*}[!htb]
\centering
{
\small
\begin{tabular}{p{2.8cm}|p{1.4cm}|p{1.7cm}|p{1.4cm}|p{1.4cm}|p{2.3cm}}
\hline
 & H2O & LightAutoML & AutoGluon & FEDOT & \textbf{AutoIntent (ours)} \\
\hline
\textit{Approach} & TAML \& Word2Vec & TAML \& emb. & Encoder fine-tuning & TAML \& TF-IDF & \textbf{Embeddings} \\
\textit{Scarce train data} & $\times$ & Has small data modes & $\times$ & Adaptable & \textbf{Adapted for small datasets}\\
\textit{Custom search space} & Flexible API & HP presets & \checkmark & HP presets & \textbf{Presets \& customizble configs} \\
\textit{Experiments tracking} & H2O Flow integration & $\times$ & $\times$ & $\times$ & \textbf{W\&B\textsuperscript{*}, tensorboard\textsuperscript{*}, codecarbon\textsuperscript{*}}. \\
\textit{Embedding prompting} & $\times$ & $\times$ & $\times$ & $\times$ & \checkmark \\
\textit{OOS detection} & $\times$ & $\times$ & $\times$ & $\times$ & \checkmark\\
\textit{Multi-label} & $\times$ & \checkmark & $\times$ & $\times$ & \checkmark\\
\hline
\end{tabular}
}
\caption{Comparison of NLP functionality in AutoML frameworks: H2O \cite{ledell2020h2o}, LightAutoML \cite{vakhrushev2022lightautomlautomlsolutionlarge}, AutoGluon \cite{tang2024autogluonmultimodalautommsuperchargingmultimodal}, FEDOT \cite{nikitin2021automated} and ours AutoIntent. HP stands for hyperparameters, TAML stands for tabular AutoML. \textsuperscript{*}Weights and Biases \cite{wandb}, Tensorboard \cite{tensorflow2015-whitepaper}, CodeCarbon \cite{benoit_courty_2024_11171501}.}
\label{tab:frameworks}
\end{table*}

\section{Background}
Automated machine learning, by definition, is a tool for automating routines like data splitting for validation, hyperparameter tuning, model ensembling, and model selection. AutoML is highly relevant in scenarios where machine learning tasks need to be solved by non-experts and in conjunction with no-code ML, which is sometimes called <<ML as a service>> \cite{bisong2019overview,barga2015predictive,liberty2020elastic,ledell2020h2o,carney2020teachable,acito2023predictive}. The applications include sentiment analysis, robotics, healthcare, business analysis \cite{yuan2024automated,salehin2024automl}.

Tabular AutoML focuses on feature engineering, feature selection and model ensembling \cite{feurer2022autosklearn20handsfreeautoml, vakhrushev2022lightautomlautomlsolutionlarge, erickson2020autogluontabularrobustaccurateautoml, ledell2020h2o, nikitin2021automated}. Usually, they employ classical machine learning methods like GBMs \cite{chen2016xgboost,prokhorenkova2018catboost,ke2017lightgbm} and linear models, fast hyperparameter tuning methods with budget-aware strategies \cite{akiba2019optuna}, and ensembling strategies like stacking, blending and voting. Such frameworks sometimes can supersede exploratory data analysis and extensive research with just a running preset training recipe. It is not rare to see AutoML frameworks winning machine learning contests, but the open source solutions often are not transferable to production-ready systems as the resulting pipeline is an ensemble of a numerous amount of models without clear guides for deployment.

Neural architecture search can be viewed as an automation in the field of deep learning \cite{salehin2024automl}. It emphasizes finding optimal computational graph using approaches like cell-based and hierarchical search spaces \cite{zoph2018learningtransferablearchitecturesscalable,real2019regularizedevolutionimageclassifier} or using scaling laws \cite{tan2020efficientnetrethinkingmodelscaling}. It cannot be treated as full-fledged AutoML, as it is designed to address only the model selection problem. Though, it can be a part of an AutoML pipeline \cite{autokeras}.

AutoML tools in the NLP domain primarily stand out from other AutoML by native support of text inputs \cite{tang2024autogluonmultimodalautommsuperchargingmultimodal,vakhrushev2022lightautomlautomlsolutionlarge}. This is especially important for use by non-experts, as it removes the requirement of manual tokenization and vectorization. Though, some tabular AutoML frameworks provide auxiliary tools for text feature extraction \cite{ledell2020h2o}. The next peculiarity of text AutoML frameworks is their usage of transformer-based backbones, which makes sense, as this is the state-of-the-art in the field of NLP. Note that NLP AutoML primarily focuses on simple tasks like classification and regression, ignoring text generation and named entity recognition, for instance.

In AutoML frameworks, the model selection can be implemented in three ways. The first and most straightforward is to use hyperparameter tuning tools like Optuna \cite{akiba2019optuna} and genetic algorithms \cite{feurer2022autosklearn20handsfreeautoml} with preset search spaces. Usually, these presets differ in how much time and computational resources they require to reach acceptable quality. The variety of presets is provided to cover all possible use cases and hardware settings. Another option is not to tune hyperparameters but use some generic hyperparameters that reach balance between the final quality and the generalization across different tasks. These hyperparameters can be obtained empirically \cite{tang2024autogluonmultimodalautommsuperchargingmultimodal}. The compromise between freezing hyperparameters and tuning all of them is the meta-learning \cite{Desai2022,feurer2022autosklearn20handsfreeautoml,wang2021meta,tian2022meta,huisman2021survey}, where metamodel takes a dataset as input and predicts hyperparameters.

\begin{figure*}[!htb]
    \centering
    \includegraphics[width=1.05\linewidth]{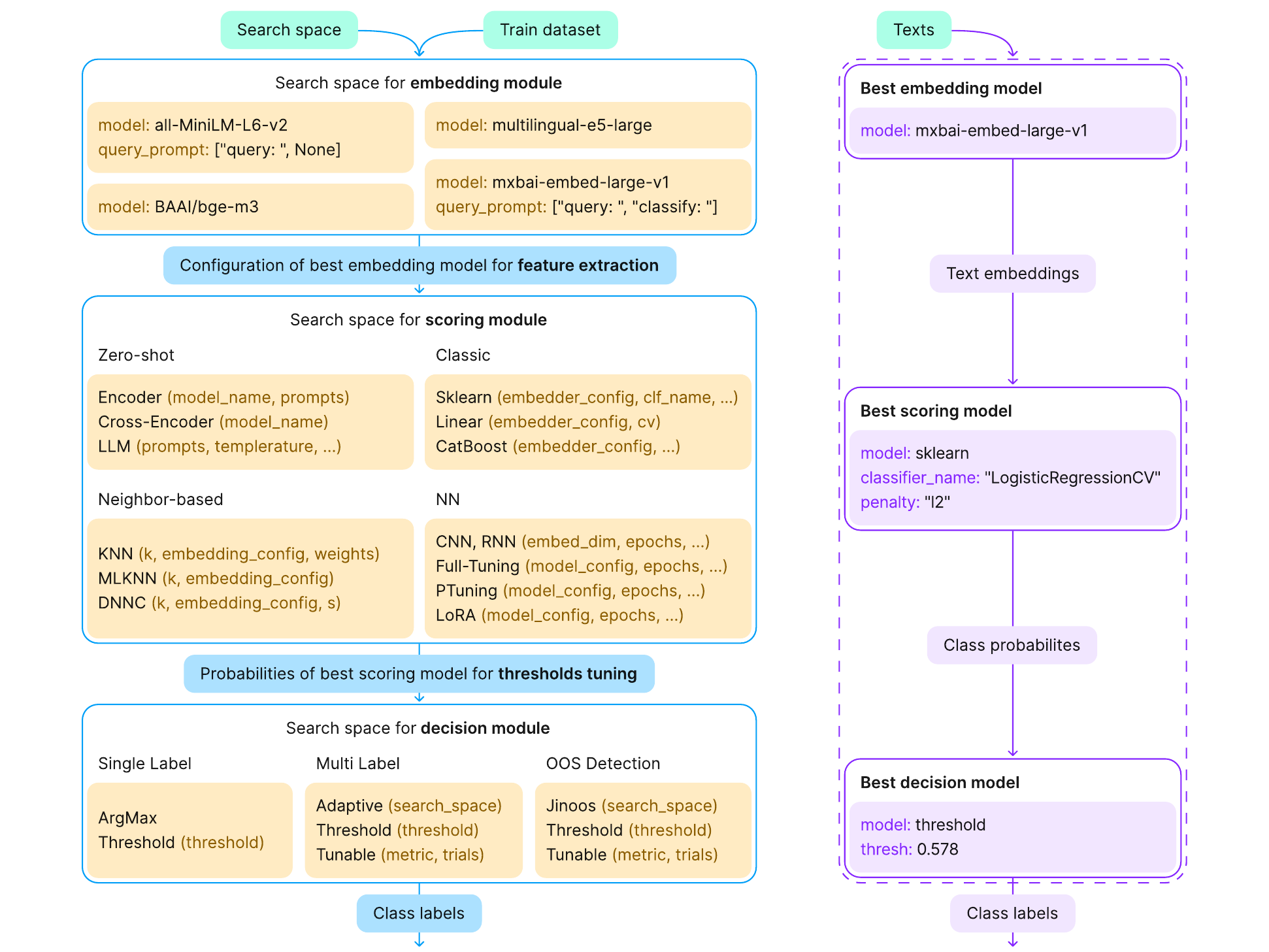}
    \caption{\textbf{(Left)} AutoIntent's three levels of hyperparameter optimization: at the module level, the embedding, scoring, and decision models are optimized sequentially; at the model level, each classification approach is tested against each other to select the best one; at the instance level, hyperparameters for each model is tuned individually with optuna samplers. \textbf{(Right)} Inference pipeline as a result of AutoIntent's hyperparameter optimization.}
    \label{fig:autointent_pipeline}
\end{figure*}

\section{AutoIntent}

\subsection{Design Principles}

The design of AutoIntent is guided by several key principles to ensure practical usability and maintainability (see Table \ref{tab:frameworks}). It features a \textbf{modular architecture} with a clear separation of concerns, adhering to \textbf{software engineering best practices} like type checking, auto-testing, and comprehensive documentation. The framework offers \textbf{model diversity}, supporting both high-performance deep learning models and efficient classical ML models that operate on pre-computed transformer embeddings. This \textbf{embedding-centric design} leverages the HuggingFace model repository and eliminates complex feature engineering. For usability, AutoIntent provides \textbf{flexible optimization strategies} (from presets to custom search spaces), multi-label classification, out-of-scope (OOS) detection, and few-shot learning.

\subsection{Separation of Concerns}

AutoIntent defines a \textit{scoring module} as a section of the text classification pipeline that outputs class probabilities, establishing a clear separation from the decision module, which makes the final prediction by applying thresholds. This separation enhances modularity and flexibility, allowing a single scoring model's outputs to be reused with various decision strategies without re-computation, which is highly efficient for experimentation.

\begin{table*}[!ht]
  \centering
  \begin{tabular}{l|c|ccccc|c}
  \hline
  preset & duration & banking77 & hwu64 & massive & minds14 & snips & avg \\
  \hline
  \multicolumn{8}{c}{\textit{Baselines}} \\
  \hline
  AutoGluon (best)       & -- &  6.98 & 12.64 & 21.39 & 85.19 & 96.00 & 44.44 \\
  AutoGluon (high)       & -- & \textbf{92.60} & 90.80 & \textbf{89.22} & 95.37 & 98.86 & 93.37 \\
  AutoGluon (medium)     & 461 & 92.40 & \textbf{91.17} & 87.13 & 92.59 & 98.86 & 92.43 \\
  LightAutoML            & 430 & 53.31 & 77.85 & 47.41 & 72.22 & 98.38 & 69.83 \\
  h2o & 257 & 75.32 & 77.32 & 75.30 & 76.85 & 98.36 & 80.63 \\
  \hline
  \multicolumn{8}{c}{\textit{AutoIntent Presets}} \\
  \hline
  zero-shot-transformers & \textbf{24} & 69.51 & 71.47 & 63.58 & 87.04 & 89.43 & 76.21 \\
  nn-medium & 44 & 79.95 & 70.79 & 72.75 & 75.31 & 96.74 & 79.11 \\
  nn-heavy & 47 & 78.84 & 72.96 & 73.39 & 80.86 & 97.40 & 80.69 \\
  zero-shot-openai & 27 & 76.43 & 85.04 & 80.49 & 96.30 & 96.86 & 87.02 \\
  classic-light & 136 & 92.23 & 90.83 & 87.11 & 97.53 & 98.43 & 93.23 \\
  classic-medium & 216 & 92.34 & 90.92 & 87.19 & \textbf{97.84} & \textbf{98.98} & \textbf{93.45} \\
  \hline
  \end{tabular}
  \caption{Performance comparison across different presets averaged from three runs (except H2O and AutoGluon which were launched once). \textbf{Column 1}: Baseline AutoML frameworks: AutoGluon \cite{tang2024autogluonmultimodalautommsuperchargingmultimodal} with non-HPO presets \texttt{best\_quality}, \texttt{high\_quality}, \texttt{medium\_quality}, H2O \cite{ledell2020h2o} with their word2vec, LightAutoML \cite{vakhrushev2022lightautomlautomlsolutionlarge}; and AutoIntent presets: \texttt{nn} (CNN \cite{cnn}, RNN), \texttt{zero-shot} (description-based bi- and cross-encoder, LLM prompting), \texttt{classic} (knn, logreg, random forest, catboost \cite{prokhorenkova2018catboost}). \textbf{Column 2}: Duration in seconds evaluated on minds14 (Intel(R) Xeon(R) CPU E5-2698 v4 @ 2.20GHz, single Tesla P100-SXM2-16GB). \textbf{Columns 3--7}: Accuracy on test sets.}
  \label{tab:presets}
  \end{table*}

\subsection{Embedding Module}\label{sec:embedding}

AutoIntent leverages the sentence-transformers library \cite{reimers-2019-sentence-bert}, providing access to a wide range of pre-trained transformer models from Hugging Face Hub \cite{wolf2020huggingfacestransformersstateoftheartnatural}. AutoIntent offers three strategies for embedding model selection:

\textbf{Pipeline-level optimization}. The embedding model is chosen once at the start of the pipeline to maximize efficiency. The selection is based on either retrieval metrics (e.g., NDCG) or the performance of a simple downstream classifier (e.g., logistic regression).

\textbf{Scoring-level optimization}. The embedding model is optimized individually for each candidate model during the optimization of the scoring module. This is more computationally intensive, but may yield better performance.

\textbf{Fixed embedding}. Users can specify a default embedding model to skip optimization entirely.

This flexible approach allows users to balance optimization quality and computational cost.

\subsection{Scoring Module}

AutoIntent offers a diverse set of scoring models. A key architectural feature is that all the classifiers are able to operate on pre-computed embeddings. This separates computationally intensive embedding generation from the lightweight classification step, enabling a balance between effectiveness and efficiency and allowing deployment on CPU-only systems. The available scoring modules include:

\textbf{KNN-based approaches}. These include K-Nearest Neighbors method with FAISS \cite{douze2024faiss} for efficient search, a two-stage cross-encoder re-ranking approach, and MLKNN \cite{zhang2007ml} for multi-label tasks.

\textbf{BERT-based classifiers}. Support full model fine-tuning and parameter-efficient approaches like LoRA \cite{hu2021loralowrankadaptationlarge} and P-Tuning \cite{peft}.

\textbf{Generic sklearn integration} allows use of any sklearn classifier operating on embedding vectors.

\textbf{Zero-shot methods} utilize text descriptions of classes and either measure the closeness with bi- or cross-encoder or prompt LLM by API \cite{openai_api}.

\begin{table*}[!htb]
\centering
\begin{tabular}{l|ccc}
\hline
framework & in domain accuracy & out-of-scope F1-measure \\
\hline
AutoIntent & \textbf{96.13} & \textbf{76.79} \\
AutoGluon \cite{tang2024autogluonmultimodalautommsuperchargingmultimodal} & 95.76 & 48.53 \\
H2O \cite{ledell2020h2o} & 85.22 & 40.69 \\
\hline
\end{tabular}
\caption{Performance comparison on out-of-scope detection task on CLINC150 \cite{clinc150}.}
\label{tab:oos}
\end{table*}

\subsection{Decision Module}

The Decision Module processes scores to produce final predictions, which is crucial for multi-label and OOS scenarios.

\textbf{AdaptiveDecision} \cite{hou2020fewshotlearningmultilabelintent}: A sample-specific thresholding method for multi-label classification.

\textbf{JinoosDecision} \cite{zhang2020discriminativenearestneighborfewshot}: Finds a universal threshold that balances in-domain and OOS accuracy.

\textbf{ThresholdDecision}: Uses a fixed, user-specified threshold, suitable for use within the AutoML tuning pipeline.

\textbf{TunableDecision}: Employs Optuna \cite{akiba2019optuna} to automatically find the optimal threshold by maximizing the F1 score.

\subsection{AutoML Pipeline}

AutoIntent orchestrates the optimization of all components hierarchically (Fig. \ref{fig:autointent_pipeline}), with two distinct levels of optimization. At the highest level, the pipeline performs \textit{module-level optimization}, where it sequentially optimizes the embedding, scoring, and decision modules. Each module builds upon the best model from the previous module's optimization, creating a cohesive pipeline. For instance, the scoring module utilizes features from the best embedding model, while the decision module processes probabilities from the best classifier. This greedy approach effectively prevents combinatorial explosion while maintaining strong performance.

The second level focuses on \textit{model-level optimization}, where both the model and its hyperparameters are sampled with Optuna's random sampling and Tree-structured Parzen Estimators \cite{watanabe2023treestructuredparzenestimatorunderstanding}. This includes various transformer models for the embedding module, different classification methods for the scoring module, and multiple threshold strategies for the decision module.

A crucial aspect of the pipeline is its clear distinction between tuning (non-gradient optimization of hyperparameters) and training (gradient optimization of model weights, if applicable). AutoIntent implements careful data handling with separate data subsets for training weights and validating hyperparameter configurations, and strategies to prevent target leakage. For cross-validation, it uses out-of-fold predictions to train stacked models (as we can view the whole three-stage pipeline with embedding, scoring and decision nodes as stacked models). The optimization is configured via a dictionary-like search space, and the final optimized pipeline can be saved and used later with simple \texttt{save}, \texttt{load}, and \texttt{predict} methods.

\begin{figure*}[!htb]
  \centering
  \includegraphics[width=0.95\linewidth]{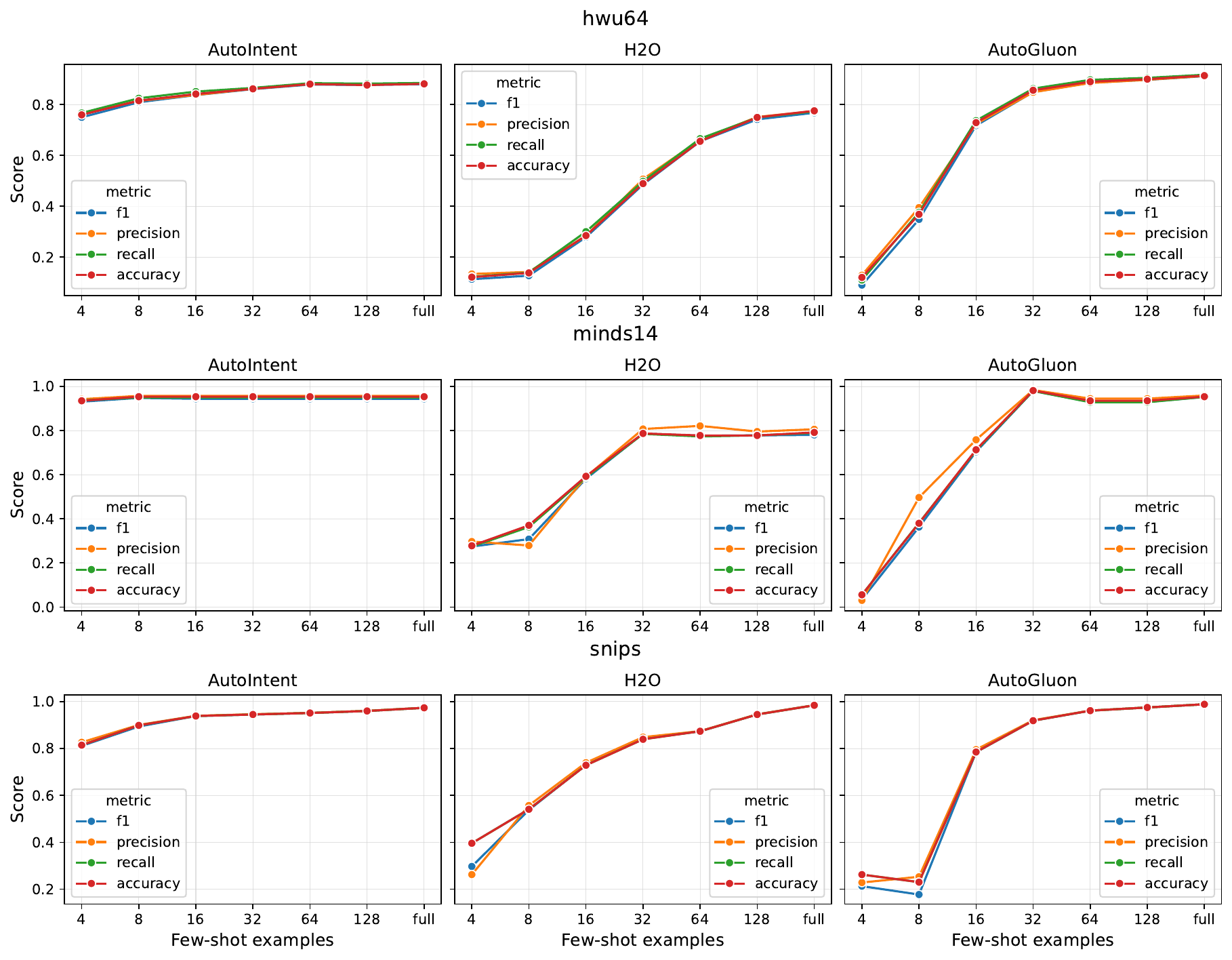}
  \caption{Performance comparison in a scenario of scarce training data. Baseline AutoML frameworks: AutoGluon \cite{tang2024autogluonmultimodalautommsuperchargingmultimodal} with non-HPO preset \texttt{medium\_quality}, H2O \cite{ledell2020h2o} with their word2vec; and AutoIntent preset \texttt{classic-light}.}
  \label{fig:fewshot}
\end{figure*}

\section{Experiments}

\subsection{Baselines}\label{sec:baselines_exps}

We compared AutoIntent against several open-source NLP AutoML frameworks: H2O \cite{ledell2020h2o}, LightAutoML (LAMA) \cite{vakhrushev2022lightautomlautomlsolutionlarge}, and AutoGluon \cite{tang2024autogluonmultimodalautommsuperchargingmultimodal}. The evaluation was conducted across five standard intent classification datasets (Table \ref{tab:presets}).

The results show that AutoGluon and AutoIntent are highly competitive, while H2O achieves moderate performance and LAMA fails on these tasks. AutoIntent provides the most affordable options in terms of balance between the quality and the computational cost. Also during the testing, we revealed several limitations in baseline frameworks:

\textbf{Hyperparameter Optimization}: AutoGluon uses a fixed training recipe; AutoGluon does support HPO presets, but they are too time and disk space consuming to test. \textbf{Feature Engineering}: H2O lacks native text features support. \textbf{Model Flexibility}: LAMA supports only three predefined transformer models. \textbf{Inference Efficiency}: AutoIntent can select lighter models for comparable performance, unlike AutoGluon's fine-tuned transformer as fixed output. \textbf{Model Variety}: Only AutoIntent provides the whole range of ML and DL models for text classification.

\subsection{OOS Detection}\label{sec:oos_detection}

We evaluated OOS capabilities on the CLINC150 dataset \cite{clinc150}. Since baselines lack native OOS support, we treated OOS as an additional class for them while using AutoIntent's built-in OOS support. The results in Table \ref{tab:oos} demonstrate AutoIntent's superiority, attributable to its dedicated confidence thresholds tuning. We utilize in-domain accuracy, because the dataset is quite balanced. Though, this is a detection task, so we consider F1-score as appropriate choice for OOS class.

\subsection{Few-shot Scenario}

We evaluated the capabilities of AutoIntent in a scenario of scarce training data. We synthetically subsampled the datasets to have only $n$ shots per class, with $n$ ranging from $4$ to $128$. The results in Figure \ref{fig:fewshot} demonstrate AutoIntent's robustness and superiority due to employing neighbor-based classification methods.

\section{Conclusion}

AutoIntent addresses the critical gap in automated machine learning for intent classification tasks, where existing AutoML frameworks lack comprehensive support for NLP-specific challenges including embedding model selection, multi-label classification, out-of-scope detection, and few-shot learning. The framework's importance stems from democratizing intent classification through end-to-end automation.

The system's novelty lies in its optimization strategy and embedding-centric design leveraging pre-computed transformer representations. AutoIntent targets NLP practitioners, conversational AI developers, and ML-as-a-service platforms.

AutoIntent operates through a three-stage pipeline (embedding, scoring, decision) with hierarchical optimization using Optuna. The embedding module selects optimal transformer models, the scoring module offers diverse classifiers.

The system was evaluated across five intent classification datasets. While demonstrating strong performance, limitations include no user studies conducted and focus on intent classification datasets. The framework is released under Apache-2.0 license to encourage community adoption.

\section{Acknowledgements}

This work was supported by the Ministry of Economic Development of the Russian Federation (agreement No. 139-15-2025-013, dated June 20, 2025, subsidy identifier 000000C313925P4B0002).

\bibliography{custom}

\begin{thebibliography}{55}
\providecommand{\natexlab}[1]{#1}

\bibitem[{Abadi et~al.(2015)Abadi, Agarwal, Barham, Brevdo, Chen, Citro,
  Corrado, Davis, Dean, Devin, Ghemawat, Goodfellow, Harp, Irving, Isard, Jia,
  Jozefowicz, Kaiser, Kudlur, Levenberg, Man\'{e}, Monga, Moore, Murray, Olah,
  Schuster, Shlens, Steiner, Sutskever, Talwar, Tucker, Vanhoucke, Vasudevan,
  Vi\'{e}gas, Vinyals, Warden, Wattenberg, Wicke, Yu, and
  Zheng}]{tensorflow2015-whitepaper}
Mart\'{i}n Abadi, Ashish Agarwal, Paul Barham, Eugene Brevdo, Zhifeng Chen,
  Craig Citro, Greg~S. Corrado, Andy Davis, Jeffrey Dean, Matthieu Devin,
  Sanjay Ghemawat, Ian Goodfellow, Andrew Harp, Geoffrey Irving, Michael Isard,
  Yangqing Jia, Rafal Jozefowicz, Lukasz Kaiser, Manjunath Kudlur, and 21
  others. 2015.
\newblock \href {https://www.tensorflow.org/} {{TensorFlow}: Large-scale
  machine learning on heterogeneous systems}.
\newblock Software available from tensorflow.org.

\bibitem[{Acito(2023)}]{acito2023predictive}
Frank Acito. 2023.
\newblock Predictive analytics with knime.
\newblock \emph{Analytics for citizen data scientists. Switzerland: Springer}.

\bibitem[{Akiba et~al.(2019)Akiba, Sano, Yanase, Ohta, and
  Koyama}]{akiba2019optuna}
Takuya Akiba, Shotaro Sano, Toshihiko Yanase, Takeru Ohta, and Masanori Koyama.
  2019.
\newblock Optuna: A next-generation hyperparameter optimization framework.
\newblock In \emph{Proceedings of the 25th ACM SIGKDD international conference
  on knowledge discovery \& data mining}, pages 2623--2631.

\bibitem[{Baratchi et~al.(2024)Baratchi, Wang, Limmer, van Rijn, Hoos,
  B{\"a}ck, and Olhofer}]{baratchi2024automated}
Mitra Baratchi, Can Wang, Steffen Limmer, Jan~N van Rijn, Holger Hoos, Thomas
  B{\"a}ck, and Markus Olhofer. 2024.
\newblock Automated machine learning: past, present and future.
\newblock \emph{Artificial intelligence review}, 57(5):122.

\bibitem[{Barga et~al.(2015)Barga, Fontama, Tok, and
  Cabrera-Cordon}]{barga2015predictive}
Roger Barga, Valentine Fontama, Wee~Hyong Tok, and Luis Cabrera-Cordon. 2015.
\newblock \emph{Predictive analytics with Microsoft Azure machine learning}.
\newblock Springer.

\bibitem[{Biewald(2020)}]{wandb}
Lukas Biewald. 2020.
\newblock \href {https://www.wandb.com/} {Experiment tracking with weights and
  biases}.
\newblock Software available from wandb.com.

\bibitem[{Bisong(2019)}]{bisong2019overview}
Ekaba Bisong. 2019.
\newblock An overview of google cloud platform services.
\newblock \emph{Building Machine learning and deep learning models on google
  cloud platform: a comprehensive guide for beginners}, pages 7--10.

\bibitem[{Carney et~al.(2020)Carney, Webster, Alvarado, Phillips, Howell,
  Griffith, Jongejan, Pitaru, and Chen}]{carney2020teachable}
Michelle Carney, Barron Webster, Irene Alvarado, Kyle Phillips, Noura Howell,
  Jordan Griffith, Jonas Jongejan, Amit Pitaru, and Alexander Chen. 2020.
\newblock Teachable machine: Approachable web-based tool for exploring machine
  learning classification.
\newblock In \emph{Extended abstracts of the 2020 CHI conference on human
  factors in computing systems}, pages 1--8.

\bibitem[{Casanueva et~al.(2020)Casanueva, Temčinas, Gerz, Henderson, and
  Vulić}]{banking77}
Iñigo Casanueva, Tadas Temčinas, Daniela Gerz, Matthew Henderson, and Ivan
  Vulić. 2020.
\newblock \href {https://arxiv.org/abs/2003.04807} {Efficient intent detection
  with dual sentence encoders}.
\newblock \emph{Preprint}, arXiv:2003.04807.

\bibitem[{Chen and Guestrin(2016)}]{chen2016xgboost}
Tianqi Chen and Carlos Guestrin. 2016.
\newblock Xgboost: A scalable tree boosting system.
\newblock In \emph{Proceedings of the 22nd acm sigkdd international conference
  on knowledge discovery and data mining}, pages 785--794.

\bibitem[{Coucke et~al.(2018)Coucke, Saade, Ball, Bluche, Caulier, Leroy,
  Doumouro, Gisselbrecht, Caltagirone, Lavril, Primet, and Dureau}]{snips}
Alice Coucke, Alaa Saade, Adrien Ball, Théodore Bluche, Alexandre Caulier,
  David Leroy, Clément Doumouro, Thibault Gisselbrecht, Francesco Caltagirone,
  Thibaut Lavril, Maël Primet, and Joseph Dureau. 2018.
\newblock \href {https://arxiv.org/abs/1805.10190} {Snips voice platform: an
  embedded spoken language understanding system for private-by-design voice
  interfaces}.
\newblock \emph{Preprint}, arXiv:1805.10190.

\bibitem[{Courty et~al.(2024)Courty, Schmidt, Luccioni, Goyal-Kamal,
  MarionCoutarel, Feld, Lecourt, LiamConnell, Saboni, Inimaz, supatomic,
  Léval, Blanche, Cruveiller, ouminasara, Zhao, Joshi, Bogroff, de~Lavoreille,
  Laskaris, Abati, Blank, Wang, Catovic, Alencon, Michał Stęchły, Bauer,
  de~Araújo, JPW, and MinervaBooks}]{benoit_courty_2024_11171501}
Benoit Courty, Victor Schmidt, Sasha Luccioni, Goyal-Kamal, MarionCoutarel,
  Boris Feld, Jérémy Lecourt, LiamConnell, Amine Saboni, Inimaz, supatomic,
  Mathilde Léval, Luis Blanche, Alexis Cruveiller, ouminasara, Franklin Zhao,
  Aditya Joshi, Alexis Bogroff, Hugues de~Lavoreille, and 11 others. 2024.
\newblock \href {https://doi.org/10.5281/zenodo.11171501} {mlco2/codecarbon:
  v2.4.1}.

\bibitem[{Desai et~al.(2022)Desai, Shah, Kothari, Surve, and
  Shekokar}]{Desai2022}
Rushil Desai, Aditya Shah, Shourya Kothari, Aishwarya Surve, and Narendra
  Shekokar. 2022.
\newblock \href {https://doi.org/10.14569/IJACSA.2022.0130988} {Textbrew:
  Automated model selection and hyperparameter optimization for text
  classification}.
\newblock \emph{International Journal of Advanced Computer Science and
  Applications}, 13(9).

\bibitem[{Devlin et~al.(2019)Devlin, Chang, Lee, and
  Toutanova}]{devlin2019bertpretrainingdeepbidirectional}
Jacob Devlin, Ming-Wei Chang, Kenton Lee, and Kristina Toutanova. 2019.
\newblock \href {https://arxiv.org/abs/1810.04805} {Bert: Pre-training of deep
  bidirectional transformers for language understanding}.
\newblock \emph{Preprint}, arXiv:1810.04805.

\bibitem[{Douze et~al.(2024)Douze, Guzhva, Deng, Johnson, Szilvasy, Mazar{\'e},
  Lomeli, Hosseini, and J{\'e}gou}]{douze2024faiss}
Matthijs Douze, Alexandr Guzhva, Chengqi Deng, Jeff Johnson, Gergely Szilvasy,
  Pierre-Emmanuel Mazar{\'e}, Maria Lomeli, Lucas Hosseini, and Herv{\'e}
  J{\'e}gou. 2024.
\newblock The faiss library.
\newblock \emph{arXiv preprint arXiv:2401.08281}.

\bibitem[{Enevoldsen et~al.(2025)Enevoldsen, Chung, Kerboua, Kardos, Mathur,
  Stap, Gala, Siblini, Krzemiński, Winata, Sturua, Utpala, Ciancone,
  Schaeffer, Sequeira, Misra, Dhakal, Rystrøm, Solomatin, Çağatan, Kundu,
  Bernstorff, Xiao, Sukhlecha, Pahwa, Poświata, Gv, Ashraf, Auras, Plüster,
  Harries, Magne, Mohr, Hendriksen, Zhu, Gisserot-Boukhlef, Aarsen, Kostkan,
  Wojtasik, Lee, Šuppa, Zhang, Rocca, Hamdy, Michail, Yang, Faysse, Vatolin,
  Thakur, Dey, Vasani, Chitale, Tedeschi, Tai, Snegirev, Günther, Xia, Shi,
  Lù, Clive, Krishnakumar, Maksimova, Wehrli, Tikhonova, Panchal, Abramov,
  Ostendorff, Liu, Clematide, Miranda, Fenogenova, Song, Safi, Li, Borghini,
  Cassano, Su, Lin, Yen, Hansen, Hooker, Xiao, Adlakha, Weller, Reddy, and
  Muennighoff}]{enevoldsen_mmteb_2025}
Kenneth Enevoldsen, Isaac Chung, Imene Kerboua, Márton Kardos, Ashwin Mathur,
  David Stap, Jay Gala, Wissam Siblini, Dominik Krzemiński, Genta~Indra
  Winata, Saba Sturua, Saiteja Utpala, Mathieu Ciancone, Marion Schaeffer,
  Gabriel Sequeira, Diganta Misra, Shreeya Dhakal, Jonathan Rystrøm, Roman
  Solomatin, and 67 others. 2025.
\newblock \href {https://arxiv.org/abs/2502.13595v2} {{{MMTEB}}: {{Massive
  Multilingual Text Embedding Benchmark}}}.

\bibitem[{Erickson et~al.(2020)Erickson, Mueller, Shirkov, Zhang, Larroy, Li,
  and Smola}]{erickson2020autogluontabularrobustaccurateautoml}
Nick Erickson, Jonas Mueller, Alexander Shirkov, Hang Zhang, Pedro Larroy,
  Mu~Li, and Alexander Smola. 2020.
\newblock \href {https://arxiv.org/abs/2003.06505} {Autogluon-tabular: Robust
  and accurate automl for structured data}.
\newblock \emph{Preprint}, arXiv:2003.06505.

\bibitem[{Feurer et~al.(2022)Feurer, Eggensperger, Falkner, Lindauer, and
  Hutter}]{feurer2022autosklearn20handsfreeautoml}
Matthias Feurer, Katharina Eggensperger, Stefan Falkner, Marius Lindauer, and
  Frank Hutter. 2022.
\newblock \href {https://arxiv.org/abs/2007.04074} {Auto-sklearn 2.0:
  Hands-free automl via meta-learning}.
\newblock \emph{Preprint}, arXiv:2007.04074.

\bibitem[{FitzGerald et~al.(2022)FitzGerald, Hench, Peris, Mackie, Rottmann,
  Sanchez, Nash, Urbach, Kakarala, Singh, Ranganath, Crist, Britan, Leeuwis,
  Tur, and Natarajan}]{massive}
Jack FitzGerald, Christopher Hench, Charith Peris, Scott Mackie, Kay Rottmann,
  Ana Sanchez, Aaron Nash, Liam Urbach, Vishesh Kakarala, Richa Singh, Swetha
  Ranganath, Laurie Crist, Misha Britan, Wouter Leeuwis, Gokhan Tur, and Prem
  Natarajan. 2022.
\newblock \href {https://arxiv.org/abs/2204.08582} {Massive: A 1m-example
  multilingual natural language understanding dataset with 51
  typologically-diverse languages}.
\newblock \emph{Preprint}, arXiv:2204.08582.

\bibitem[{Gerz et~al.(2021)Gerz, Su, Kusztos, Mondal, Lis, Singhal, Mrksic,
  Wen, and Vulic}]{minds14}
Daniela Gerz, Pei{-}Hao Su, Razvan Kusztos, Avishek Mondal, Michal Lis, Eshan
  Singhal, Nikola Mrksic, Tsung{-}Hsien Wen, and Ivan Vulic. 2021.
\newblock \href {https://arxiv.org/abs/2104.08524} {Multilingual and
  cross-lingual intent detection from spoken data}.
\newblock \emph{CoRR}, abs/2104.08524.

\bibitem[{Hou et~al.(2020)Hou, Lai, Wu, Che, and
  Liu}]{hou2020fewshotlearningmultilabelintent}
Yutai Hou, Yongkui Lai, Yushan Wu, Wanxiang Che, and Ting Liu. 2020.
\newblock \href {https://arxiv.org/abs/2010.05256} {Few-shot learning for
  multi-label intent detection}.
\newblock \emph{Preprint}, arXiv:2010.05256.

\bibitem[{Hu et~al.(2021)Hu, Shen, Wallis, Allen-Zhu, Li, Wang, Wang, and
  Chen}]{hu2021loralowrankadaptationlarge}
Edward~J. Hu, Yelong Shen, Phillip Wallis, Zeyuan Allen-Zhu, Yuanzhi Li, Shean
  Wang, Lu~Wang, and Weizhu Chen. 2021.
\newblock \href {https://arxiv.org/abs/2106.09685} {Lora: Low-rank adaptation
  of large language models}.
\newblock \emph{Preprint}, arXiv:2106.09685.

\bibitem[{Huisman et~al.(2021)Huisman, Van~Rijn, and Plaat}]{huisman2021survey}
Mike Huisman, Jan~N Van~Rijn, and Aske Plaat. 2021.
\newblock A survey of deep meta-learning.
\newblock \emph{Artificial Intelligence Review}, 54(6):4483--4541.

\bibitem[{Jin et~al.(2023)Jin, Chollet, Song, and Hu}]{autokeras}
Haifeng Jin, François Chollet, Qingquan Song, and Xia Hu. 2023.
\newblock \href {http://jmlr.org/papers/v24/20-1355.html} {Autokeras: An automl
  library for deep learning}.
\newblock \emph{Journal of Machine Learning Research}, 24(6):1--6.

\bibitem[{Ke et~al.(2017)Ke, Meng, Finley, Wang, Chen, Ma, Ye, and
  Liu}]{ke2017lightgbm}
Guolin Ke, Qi~Meng, Thomas Finley, Taifeng Wang, Wei Chen, Weidong Ma, Qiwei
  Ye, and Tie-Yan Liu. 2017.
\newblock Lightgbm: A highly efficient gradient boosting decision tree.
\newblock \emph{Advances in neural information processing systems}, 30.

\bibitem[{Kim(2014)}]{cnn}
Yoon Kim. 2014.
\newblock \href {https://arxiv.org/abs/1408.5882} {Convolutional neural
  networks for sentence classification}.
\newblock \emph{Preprint}, arXiv:1408.5882.

\bibitem[{Larson et~al.(2019)Larson, Mahendran, Peper, Clarke, Lee, Hill,
  Kummerfeld, Leach, Laurenzano, Tang, and Mars}]{clinc150}
Stefan Larson, Anish Mahendran, Joseph~J. Peper, Christopher Clarke, Andrew
  Lee, Parker Hill, Jonathan~K. Kummerfeld, Kevin Leach, Michael~A. Laurenzano,
  Lingjia Tang, and Jason Mars. 2019.
\newblock \href {https://doi.org/10.18653/v1/D19-1131} {An evaluation dataset
  for intent classification and out-of-scope prediction}.
\newblock In \emph{Proceedings of the 2019 Conference on Empirical Methods in
  Natural Language Processing and the 9th International Joint Conference on
  Natural Language Processing (EMNLP-IJCNLP)}, pages 1311--1316, Hong Kong,
  China. Association for Computational Linguistics.

\bibitem[{LeDell and Poirier(2020)}]{ledell2020h2o}
Erin LeDell and Sebastien Poirier. 2020.
\newblock H2o automl: Scalable automatic machine learning.
\newblock In \emph{Proceedings of the AutoML Workshop at ICML}, volume 2020,
  page~24.

\bibitem[{Lee et~al.(2024)Lee, Shakir, Koenig, and Lipp}]{emb2024mxbai}
Sean Lee, Aamir Shakir, Darius Koenig, and Julius Lipp. 2024.
\newblock \href {https://www.mixedbread.ai/blog/mxbai-embed-large-v1} {Open
  source strikes bread - new fluffy embeddings model}.

\bibitem[{Li and Li(2023)}]{li2023angle}
Xianming Li and Jing Li. 2023.
\newblock Angle-optimized text embeddings.
\newblock \emph{arXiv preprint arXiv:2309.12871}.

\bibitem[{Liberty et~al.(2020)Liberty, Karnin, Xiang, Rouesnel, Coskun,
  Nallapati, Delgado, Sadoughi, Astashonok, Das et~al.}]{liberty2020elastic}
Edo Liberty, Zohar Karnin, Bing Xiang, Laurence Rouesnel, Baris Coskun, Ramesh
  Nallapati, Julio Delgado, Amir Sadoughi, Yury Astashonok, Piali Das, and 1
  others. 2020.
\newblock Elastic machine learning algorithms in amazon sagemaker.
\newblock In \emph{Proceedings of the 2020 ACM SIGMOD International Conference
  on Management of Data}, pages 731--737.

\bibitem[{Liu et~al.(2019)Liu, Eshghi, Swietojanski, and Rieser}]{hwu64}
Xingkun Liu, Arash Eshghi, Pawel Swietojanski, and Verena Rieser. 2019.
\newblock \href {https://arxiv.org/abs/1903.05566} {Benchmarking natural
  language understanding services for building conversational agents}.
\newblock \emph{Preprint}, arXiv:1903.05566.

\bibitem[{Mangrulkar et~al.(2022)Mangrulkar, Gugger, Debut, Belkada, Paul, and
  Bossan}]{peft}
Sourab Mangrulkar, Sylvain Gugger, Lysandre Debut, Younes Belkada, Sayak Paul,
  and Benjamin Bossan. 2022.
\newblock Peft: State-of-the-art parameter-efficient fine-tuning methods.
\newblock \url{https://github.com/huggingface/peft}.

\bibitem[{Muennighoff et~al.(2023)Muennighoff, Tazi, Magne, and
  Reimers}]{muennighoff_mteb_2023}
Niklas Muennighoff, Nouamane Tazi, Loïc Magne, and Nils Reimers. 2023.
\newblock \href {https://doi.org/10.48550/arXiv.2210.07316} {{{MTEB}}:
  {{Massive Text Embedding Benchmark}}}.
\newblock \emph{Preprint}, arXiv:2210.07316.

\bibitem[{Nikitin et~al.(2021)Nikitin, Vychuzhanin, Sarafanov, Polonskaia,
  Revin, Barabanova, Maximov, Kalyuzhnaya, and
  Boukhanovsky}]{nikitin2021automated}
Nikolay~O. Nikitin, Pavel Vychuzhanin, Mikhail Sarafanov, Iana~S. Polonskaia,
  Ilia Revin, Irina~V. Barabanova, Gleb Maximov, Anna~V. Kalyuzhnaya, and
  Alexander Boukhanovsky. 2021.
\newblock \href {https://doi.org/10.1016/j.future.2021.08.022} {Automated
  evolutionary approach for the design of composite machine learning
  pipelines}.
\newblock \emph{Future Generation Computer Systems}.

\bibitem[{OpenAI(2023)}]{openai_api}
OpenAI. 2023.
\newblock Openai api.
\newblock \url{https://openai.com/api/}.
\newblock Accessed: 26 may 2025.

\bibitem[{Pedregosa et~al.(2018)Pedregosa, Varoquaux, Gramfort, Michel,
  Thirion, Grisel, Blondel, Müller, Nothman, Louppe, Prettenhofer, Weiss,
  Dubourg, Vanderplas, Passos, Cournapeau, Brucher, Perrot, and Édouard
  Duchesnay}]{pedregosa2018scikitlearnmachinelearningpython}
Fabian Pedregosa, Gaël Varoquaux, Alexandre Gramfort, Vincent Michel, Bertrand
  Thirion, Olivier Grisel, Mathieu Blondel, Andreas Müller, Joel Nothman,
  Gilles Louppe, Peter Prettenhofer, Ron Weiss, Vincent Dubourg, Jake
  Vanderplas, Alexandre Passos, David Cournapeau, Matthieu Brucher, Matthieu
  Perrot, and Édouard Duchesnay. 2018.
\newblock \href {https://arxiv.org/abs/1201.0490} {Scikit-learn: Machine
  learning in python}.
\newblock \emph{Preprint}, arXiv:1201.0490.

\bibitem[{Prokhorenkova et~al.(2018)Prokhorenkova, Gusev, Vorobev, Dorogush,
  and Gulin}]{prokhorenkova2018catboost}
Liudmila Prokhorenkova, Gleb Gusev, Aleksandr Vorobev, Anna~Veronika Dorogush,
  and Andrey Gulin. 2018.
\newblock Catboost: unbiased boosting with categorical features.
\newblock \emph{Advances in neural information processing systems}, 31.

\bibitem[{Real et~al.(2019)Real, Aggarwal, Huang, and
  Le}]{real2019regularizedevolutionimageclassifier}
Esteban Real, Alok Aggarwal, Yanping Huang, and Quoc~V Le. 2019.
\newblock \href {https://arxiv.org/abs/1802.01548} {Regularized evolution for
  image classifier architecture search}.
\newblock \emph{Preprint}, arXiv:1802.01548.

\bibitem[{Reimers and Gurevych(2019)}]{reimers-2019-sentence-bert}
Nils Reimers and Iryna Gurevych. 2019.
\newblock \href {https://arxiv.org/abs/1908.10084} {Sentence-bert: Sentence
  embeddings using siamese bert-networks}.
\newblock In \emph{Proceedings of the 2019 Conference on Empirical Methods in
  Natural Language Processing}. Association for Computational Linguistics.

\bibitem[{Salehin et~al.(2024)Salehin, Islam, Saha, Noman, Tuni, Hasan, and
  Baten}]{salehin2024automl}
Imrus Salehin, Md~Shamiul Islam, Pritom Saha, SM~Noman, Azra Tuni, Md~Mehedi
  Hasan, and Md~Abu Baten. 2024.
\newblock Automl: A systematic review on automated machine learning with neural
  architecture search.
\newblock \emph{Journal of Information and Intelligence}, 2(1):52--81.

\bibitem[{Taha et~al.(2024)Taha, Yoo, Yeun, Homouz, and Taha}]{TAHA2024100664}
Kamal Taha, Paul~D. Yoo, Chan Yeun, Dirar Homouz, and Aya Taha. 2024.
\newblock \href {https://doi.org/10.1016/j.cosrev.2024.100664} {A comprehensive
  survey of text classification techniques and their research applications:
  Observational and experimental insights}.
\newblock \emph{Computer Science Review}, 54:100664.

\bibitem[{Tan and Le(2020)}]{tan2020efficientnetrethinkingmodelscaling}
Mingxing Tan and Quoc~V. Le. 2020.
\newblock \href {https://arxiv.org/abs/1905.11946} {Efficientnet: Rethinking
  model scaling for convolutional neural networks}.
\newblock \emph{Preprint}, arXiv:1905.11946.

\bibitem[{Tang et~al.(2024)Tang, Fang, Zhou, Yang, Zhong, Hu, Kirchhoff, and
  Karypis}]{tang2024autogluonmultimodalautommsuperchargingmultimodal}
Zhiqiang Tang, Haoyang Fang, Su~Zhou, Taojiannan Yang, Zihan Zhong, Tony Hu,
  Katrin Kirchhoff, and George Karypis. 2024.
\newblock \href {https://arxiv.org/abs/2404.16233} {Autogluon-multimodal
  (automm): Supercharging multimodal automl with foundation models}.
\newblock \emph{Preprint}, arXiv:2404.16233.

\bibitem[{Tian et~al.(2022)Tian, Zhao, and Huang}]{tian2022meta}
Yingjie Tian, Xiaoxi Zhao, and Wei Huang. 2022.
\newblock Meta-learning approaches for learning-to-learn in deep learning: A
  survey.
\newblock \emph{Neurocomputing}, 494:203--223.

\bibitem[{Vakhrushev et~al.(2022)Vakhrushev, Ryzhkov, Savchenko, Simakov,
  Damdinov, and Tuzhilin}]{vakhrushev2022lightautomlautomlsolutionlarge}
Anton Vakhrushev, Alexander Ryzhkov, Maxim Savchenko, Dmitry Simakov, Rinchin
  Damdinov, and Alexander Tuzhilin. 2022.
\newblock \href {https://arxiv.org/abs/2109.01528} {Lightautoml: Automl
  solution for a large financial services ecosystem}.
\newblock \emph{Preprint}, arXiv:2109.01528.

\bibitem[{Vaswani et~al.(2023)Vaswani, Shazeer, Parmar, Uszkoreit, Jones,
  Gomez, Kaiser, and Polosukhin}]{vaswani2023attentionneed}
Ashish Vaswani, Noam Shazeer, Niki Parmar, Jakob Uszkoreit, Llion Jones,
  Aidan~N. Gomez, Lukasz Kaiser, and Illia Polosukhin. 2023.
\newblock \href {https://arxiv.org/abs/1706.03762} {Attention is all you need}.
\newblock \emph{Preprint}, arXiv:1706.03762.

\bibitem[{Wang(2021)}]{wang2021meta}
Jane~X Wang. 2021.
\newblock Meta-learning in natural and artificial intelligence.
\newblock \emph{Current Opinion in Behavioral Sciences}, 38:90--95.

\bibitem[{Watanabe(2023)}]{watanabe2023treestructuredparzenestimatorunderstanding}
Shuhei Watanabe. 2023.
\newblock \href {https://arxiv.org/abs/2304.11127} {Tree-structured parzen
  estimator: Understanding its algorithm components and their roles for better
  empirical performance}.
\newblock \emph{Preprint}, arXiv:2304.11127.

\bibitem[{Weld et~al.(2021)Weld, Huang, Long, Poon, and
  Han}]{weld2021surveyjointintentdetection}
H.~Weld, X.~Huang, S.~Long, J.~Poon, and S.~C. Han. 2021.
\newblock \href {https://arxiv.org/abs/2101.08091} {A survey of joint intent
  detection and slot-filling models in natural language understanding}.
\newblock \emph{Preprint}, arXiv:2101.08091.

\bibitem[{Wolf et~al.(2020)Wolf, Debut, Sanh, Chaumond, Delangue, Moi, Cistac,
  Rault, Louf, Funtowicz, Davison, Shleifer, von Platen, Ma, Jernite, Plu, Xu,
  Scao, Gugger, Drame, Lhoest, and
  Rush}]{wolf2020huggingfacestransformersstateoftheartnatural}
Thomas Wolf, Lysandre Debut, Victor Sanh, Julien Chaumond, Clement Delangue,
  Anthony Moi, Pierric Cistac, Tim Rault, Rémi Louf, Morgan Funtowicz, Joe
  Davison, Sam Shleifer, Patrick von Platen, Clara Ma, Yacine Jernite, Julien
  Plu, Canwen Xu, Teven~Le Scao, Sylvain Gugger, and 3 others. 2020.
\newblock \href {https://arxiv.org/abs/1910.03771} {Huggingface's transformers:
  State-of-the-art natural language processing}.
\newblock \emph{Preprint}, arXiv:1910.03771.

\bibitem[{Yuan et~al.(2024)Yuan, Yu, Xie, Liu, and Sun}]{yuan2024automated}
Han Yuan, Kunyu Yu, Feng Xie, Mingxuan Liu, and Shenghuan Sun. 2024.
\newblock Automated machine learning with interpretation: a systematic review
  of methodologies and applications in healthcare.
\newblock \emph{Medicine Advances}, 2(3):205--237.

\bibitem[{Zhang et~al.(2020)Zhang, Hashimoto, Liu, Wu, Wan, Yu, Socher, and
  Xiong}]{zhang2020discriminativenearestneighborfewshot}
Jian-Guo Zhang, Kazuma Hashimoto, Wenhao Liu, Chien-Sheng Wu, Yao Wan,
  Philip~S. Yu, Richard Socher, and Caiming Xiong. 2020.
\newblock \href {https://arxiv.org/abs/2010.13009} {Discriminative nearest
  neighbor few-shot intent detection by transferring natural language
  inference}.
\newblock \emph{Preprint}, arXiv:2010.13009.

\bibitem[{Zhang and Zhou(2007)}]{zhang2007ml}
Min-Ling Zhang and Zhi-Hua Zhou. 2007.
\newblock Ml-knn: A lazy learning approach to multi-label learning.
\newblock \emph{Pattern recognition}, 40(7):2038--2048.

\bibitem[{Zoph et~al.(2018)Zoph, Vasudevan, Shlens, and
  Le}]{zoph2018learningtransferablearchitecturesscalable}
Barret Zoph, Vijay Vasudevan, Jonathon Shlens, and Quoc~V. Le. 2018.
\newblock \href {https://arxiv.org/abs/1707.07012} {Learning transferable
  architectures for scalable image recognition}.
\newblock \emph{Preprint}, arXiv:1707.07012.

\end{thebibliography}

\appendix

\section{Computational Efficiency}

To quantify the computational requirements of different scoring modules, we conducted a comprehensive analysis using the Code Carbon library \cite{benoit_courty_2024_11171501}. This analysis measured various aspects of computational resource consumption for a single trial (training and evaluation of a single model configuration). The results, presented in Table \ref{tab:computational}, reveal significant variations in resource usage across different approaches.

\begin{table*}[ht]
\centering
\begin{tabular}{l|ccccccc}
\hline
model & emissions & runtime & energy & gpu & cpu & ram & rate \\
\hline
bert & 1.382 & 103.911 & 3.133 & 2.198 & 0.774 & 1.615e-01 & 0.014 \\
ptuning & 1.118 & 83.455 & 2.535 & 1.785 & 0.620 & 1.295e-01 & 0.014 \\
lora & 0.863 & 65.157 & 1.957 & 1.372 & 0.484 & 1.009e-01 & 0.013 \\
linear & 0.428 & 73.393 & 0.971 & 0.312 & 0.545 & 1.138e-01 & 0.006 \\
rerank & 0.270 & 29.040 & 0.613 & 0.355 & 0.213 & 4.436e-02 & 0.010 \\
dnnc & 0.122 & 10.000 & 0.276 & 0.192 & 0.070 & 1.455e-02 & 0.013 \\
rand forest & 0.073 & 11.367 & 0.166 & 0.074 & 0.080 & 1.664e-02 & 0.007 \\
knn & 0.009 & 1.281 & 0.019 & 0.014 & 0.004 & 9.044e-04 & 0.012 \\
\hline
units & grams & sec & Wh & Wh & Wh & Wh & grams/sec \\
\hline
\end{tabular}
\caption{Computational resource consumption for different scoring modules. The experiments are conducted on banking77 dataset with \texttt{mixedbread-ai/mxbai-embed-large-v1} \cite{emb2024mxbai,li2023angle}, system with AMD Ryzen 7 5800H, NVIDIA RTX 3060 Laptop. Median values of 10 trials are displayed. Embeddings were pre-computed.}
\label{tab:computational}
\end{table*}

The analysis revealed significant variations in resource efficiency across different scoring methods. KNN-based methods demonstrated exceptional efficiency, with minimal emissions and runtime, making them particularly suitable for resource-constrained environments. Logistic regression showed moderate resource consumption while maintaining high performance, representing a balanced trade-off between computational cost and effectiveness. In contrast, BERT-based methods exhibited the highest resource requirements, necessitating substantial computational infrastructure. These findings provide valuable insights for deployment scenarios with varying resource constraints and for building AutoIntent's presets.

\subsection{Embedding Module Effectiveness}
\begin{figure}[!htb]
\centering
\includegraphics[width=1\linewidth]{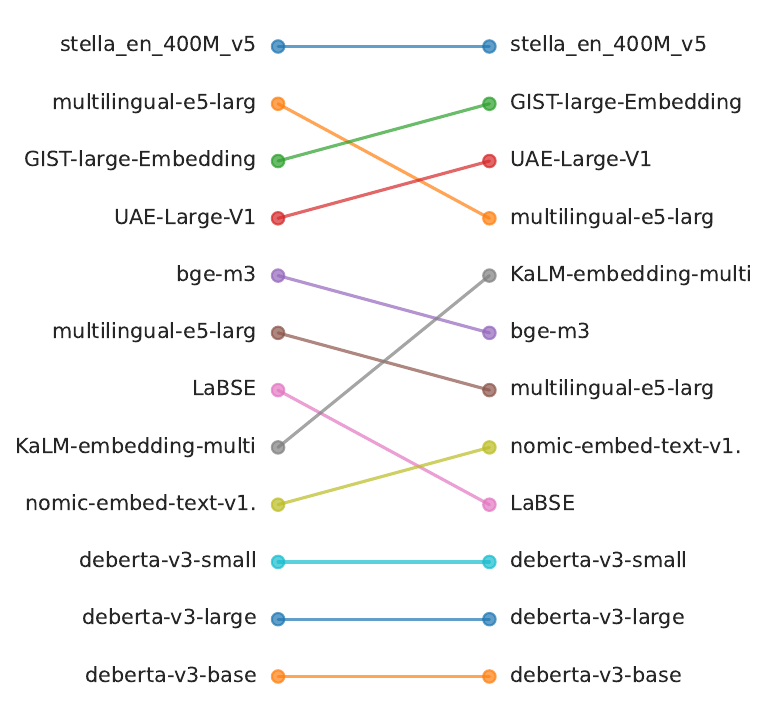}
\caption{Encoders ranking: (\textbf{Left}) precise ranking obtained via training full AutoML pipeline with only this model, (\textbf{Right}) approximate ranking based on retrieval quality (NDCG).}
\label{fig:encoder_ranking}
\end{figure}

We evaluated our retrieval-based embedding selection heuristic (optimizing NDCG) against the ground truth (final accuracy from the full pipeline), as described in Section \ref{sec:embedding}. As shown in Figure \ref{fig:encoder_ranking} and Table \ref{tab:encoder_ranking}, while the approximate ranking is imperfect, it successfully identifies the best model (\texttt{stella\_en\_400M\_v5}). This demonstrates that the heuristic effectively balances computational cost and selection quality. We have taken top models from \texttt{MTEB(eng)}\cite{muennighoff_mteb_2023, enevoldsen_mmteb_2025} leaderboard.

\begin{table}[ht]
\begin{tabular}{l|cc}
\hline
\textbf{Model} & \textbf{Accuracy} & \textbf{NDCG} \\
\hline
stella\_en\_400M\_v5 & 94.28 & 93.83 \\
multilingual-e5-\textsuperscript{1} & 93.65 & 92.97 \\
GIST-large-\textsuperscript{2} & 93.51 & 93.32 \\
UAE-Large-V1 & 92.89 & 93.25 \\
bge-m3 & 92.69 & 92.49 \\
multilingual-e5-large & 91.41 & 92.45 \\
LaBSE & 90.47 & 89.51 \\
KaLM-embedding-\textsuperscript{3} & 89.65 & 92.88 \\
nomic-embed-\textsuperscript{4} & 87.24 & 89.63 \\
deberta-v3-small & 81.15 & 67.20 \\
deberta-v3-large & 75.39 & 59.59 \\
deberta-v3-base & 75.00 & 59.28 \\
\hline
\end{tabular}
\caption{Embedding models performance averaged over hwu64 \cite{hwu64}, massive \cite{massive}, minds14 \cite{minds14}, snips \cite{snips}. \textsuperscript{1}large-unstruct, \textsuperscript{2}Embedding-v0, \textsuperscript{3}multilingual-mini-instruct-v1.5, \textsuperscript{4}text-v1.5}
\label{tab:encoder_ranking}
\end{table}

\end{document}